# A brief survey on deep belief networks and introducing a new object oriented toolbox (DeeBNet V3.0)

Mohammad Ali Keyvanrad [a], Mohammad Mehdi Homayounpour [a]

[a] *Laboratory for Intelligent Multimedia Processing (LIMP), Computer Engineering and Information Technology Department, Amirkabir University of Technology, Tehran, Iran*

Phone: +98(21)64542747

Fax: +98(21)64542700

Email: {keyvanrad, homayoun}@aut.ac.ir

Abstract. Nowadays, this is very popular to use the deep architectures in machine learning. Deep Belief Networks (DBNs) are deep architectures that use stack of Restricted Boltzmann Machines (RBM) to create a powerful generative model using training data. DBNs have many ability like feature extraction and classification that are used in many applications like image processing, speech processing and etc. This paper introduces a new object oriented MATLAB toolbox with most of abilities needed for the implementation of DBNs. In the new version, the toolbox can be used in Octave. According to the results of the experiments conducted on MNIST (image), ISOLET (speech), and 20 Newsgroups (text) datasets, it was shown that the toolbox can learn automatically a good representation of the input from unlabeled data with better discrimination between different classes. Also on all datasets, the obtained classification errors are comparable to those of state of the art classifiers. In addition, the toolbox supports different sampling methods (e.g. Gibbs, CD, PCD and our new FEPCD method), different sparsity methods (quadratic, rate distortion and our new normal method), different RBM types (generative and discriminative), using GPU, etc. The toolbox is a user-friendly open source software and is freely available on the website http://ceit.aut.ac.ir/~keyvanrad/DeeBNet%20Toolbox.html .

*Keywords: Deep Belief Network, Restricted Boltzmann Machine, Artificial Neural Network, MATLAB Toolbox, Classification, Feature Extraction, Sparse RBM*

## 1. Introduction

Since many years ago, artificial neural networks have been used in artificial intelligence applications. Pattern recognition, voice and speech analysis and natural language processing are some of these applications that use artificial neural networks. Due to some theoretical and biological reasons, deep models and architectures with many nonlinear processing layers were suggested.

These deep models have many layers and parameters that must be learnt. When the learning process is so complicated and a huge number of parameters are needed, artificial neural networks are rarely used. The problem of this number of layers is that training is time consuming and training becomes trapped at local minima. Therefore we can't achieve acceptable results. One important tool for dealing with this problem is to use DBNs (Deep Belief Network) that can create neural networks including many hidden layers [1].Deep Belief Networks can be used in classification and feature learning. Data representation is very important in machine learning. Therefore, much work has been done for feature preprocessing, feature extraction and feature learning. In feature learning, we can create a feature extraction system and then use the extracted features in classification and other applications. Using unlabeled data in high level feature extraction [2] and also increasing discrimination between extracted features are the benefits of DBN for feature learning [3].

Layers of DBN are created from Restricted Boltzmann Machine (RBM) that is a generative and undirected probabilistic model. RBMs use a hidden layer to model the probability distribution of visible variables. Indeed, we can create a DBN for hierarchical processing using stacking RBMs. Therefore most of improvements in DBNs are due to improvement in RBMs. This paper studies different developed RBM models and introduces a new MATLAB and Octave toolbox with many DBN abilities.

Hinton presented DBNs and used it in the task of digit recognition on MNIST data set [4]. He used a DBN with 784-500-500-2000-10 structure, where the first layer possesses 784 features from 28*28 MNIST digit images. The last layer is related to 10 digit labels and other three layers are hidden layers with stochastic binary neurons. Finally this paper achieved 1.25% classification error rate on MNIST test data set.

In another paper from this author [3], DBN is used as a nonlinear model for feature extraction and dimension reduction. Indeed, the DBN may be considered as a model that can generate features in its last layer with the ability to reconstruct visible data from generated features. When a general Neural Network is used with many layers, the Neural Network becomes trapped in local minima and the performance will decrease. Therefore determining the initial values for NN weights is critical.

Another paper proposed DDBN (Discriminative Deep Belief Network) is based on DBN as a new classifier [1]. This paper showed the power of DBN in using unlabeled data and also performance improvement by increasing layers (even by 50 hidden layers).

DBN applications are not limited to image processing and can be used in voice processing [5]–[8] with significant efficiency. Some toolkits have been developed and introduced to facilitate the use of DBNs in different applications. The implemented toolboxes are developed to be used for many different tasks including classification, feature extraction, data reconstructing, noise reduction, generating new data, etc. Some of these toolboxes are listed and compared in Table 1. The comparison is based on some features and characteristics including programming language, open source object oriented programming, learning method, discriminative ability, type of visible nodes, fine tuning, possibility of being used using GPUs, and documentation. As Table 1 depicts, in comparison to other DBN toolboxes, our toolbox possesses all main features as well as different types of classes. Also it is designed to be very modular, extensible and reusable.

The rest of this paper is organized as follows: in section 2, RBM and DBN are described. The introducing our new MATLAB toolbox and some experiments on MNIST and ISOLET dataset is presented in section 3. Finally, section 4 concludes the paper.

*Table 1: A brief comparison with other implemented toolboxes.*

| Toolkit Name | Progr. Lang | open source | OOP[1] | Learning method | DRBM[2] | Sparse RBM | type visible nodes | fine-tuning | GPU | User Manual |
|---|---|---|---|---|---|---|---|---|---|---|
| **deepLearn, 2014[3]** | MATLAB | ✓ | ✗ | CD1 | ✗ | ✓ | probability | ✓ | ✗ | Incomplete |
| **deep autoencoder, 2006[4]** | MATLAB | ✓ | ✗ | CD1 | ✗ | ✗ | probability | ✓ | ✗ | Incomplete |
| **matrbm, 2010[5]** | MATLAB | ✓ | ✗ | CD1, PCD | ✓ | ✗ | probability | ✗ | ✗ | Incomplete |
| **deepmat, 2014[6]** | MATLAB | ✓ | ✗ | CDk, PCD, FPCD | ✓ | ✓ | Probability, Gaussian | ✓ | ✓ | Incomplete |
| **DigitDemo, 2010[7]** | MATLAB | ✗ | ✗ | CDk, PCD, RM, PL | ✗ | ✗ | Probability | ✓ | ✗ | Incomplete |
| **DBN Toolbox, 2010[8]** | MATLAB | ✓ | ✓ | CDk | ✗ | ✗ | Probability, Gaussian | ✓ | ✗ | Incomplete |
| **DeeBNet (our toolbox)** | MATLAB | ✓ | ✓ | Gibbs, CDk, PCD, FEPCD | ✓ | ✓ | binary, probability, Gaussian | ✓ | ✓ | complete (in English), perfect (in Persian) |

---

1 Object-oriented programming
2 Discriminative Restricted Boltzmann Machine
3 Rasmus Berg Palm, https://github.com/rasmusbergpalm/DeepLearnToolbox
4 Ruslan Salakhutdinov and Geoff Hinton, http://www.cs.toronto.edu/~hinton/MatlabForSciencePaper.html
5 Andrej Karpathy, https://code.google.com/p/matrbm
6 Kyunghyun Cho, https://github.com/kyunghyuncho/deepmat
7 Benjamin M. Marlin, https://people.cs.umass.edu/~marlin/code-digitdemo.shtml
8 Drausin Wulsin, http://www.seas.upenn.edu/~wulsin

# 2. Deep Belief Networks (DBNs) and Restricted Boltzmann Machines (RBMs)

DBNs are composed of multiple layers of RBMs. RBM is a Boltzmann machine where the connections between hidden visible layers are disjointed. Also the Boltzmann machine is an undirected graphical model (or Markov Random Field). In the Following section, the RBMs and some revised version of RBMs are discussed. It is explained how DBNs are constructed using Restricted Boltzmann Machines (RBMs).

The Boltzmann Machine is a type of MRF. The Boltzmann Machine is a concurrent network with stochastic binary units. The network has a set of visible units $v \in \{0,1\}^{g_v}$ and a set of hidden units $h \in \{0,1\}^{g_h}$ where $g_v$ and $g_h$ are the number of visible units and the number of hidden units respectively (left figure in Figure 1). The energy of the joint configuration $\{\mathrm{v}, \mathrm{h}\}$ in Boltzmann machine is given as follows:

$$E(v,h) = -\frac{1}{2}v^T L v - \frac{1}{2}h^T J h - v^T W h \tag{1}$$

The bias is removed for simplicity of presentation. The term $W$ is the concurrent weights between visible and hidden units, $L$ is the concurrent weights between visible and visible units and finally $J$ is the concurrent weights between hidden and hidden units. Diagonal values of $L$ and $J$ are zero.

Since Boltzmann machines have a complicated theory and formulations, therefore Restricted Boltzmann Machines are used for simplicity. If $J = 0$ and $L = 0$, the famous RBM model is introduced (the right hand figure in Figure 1).

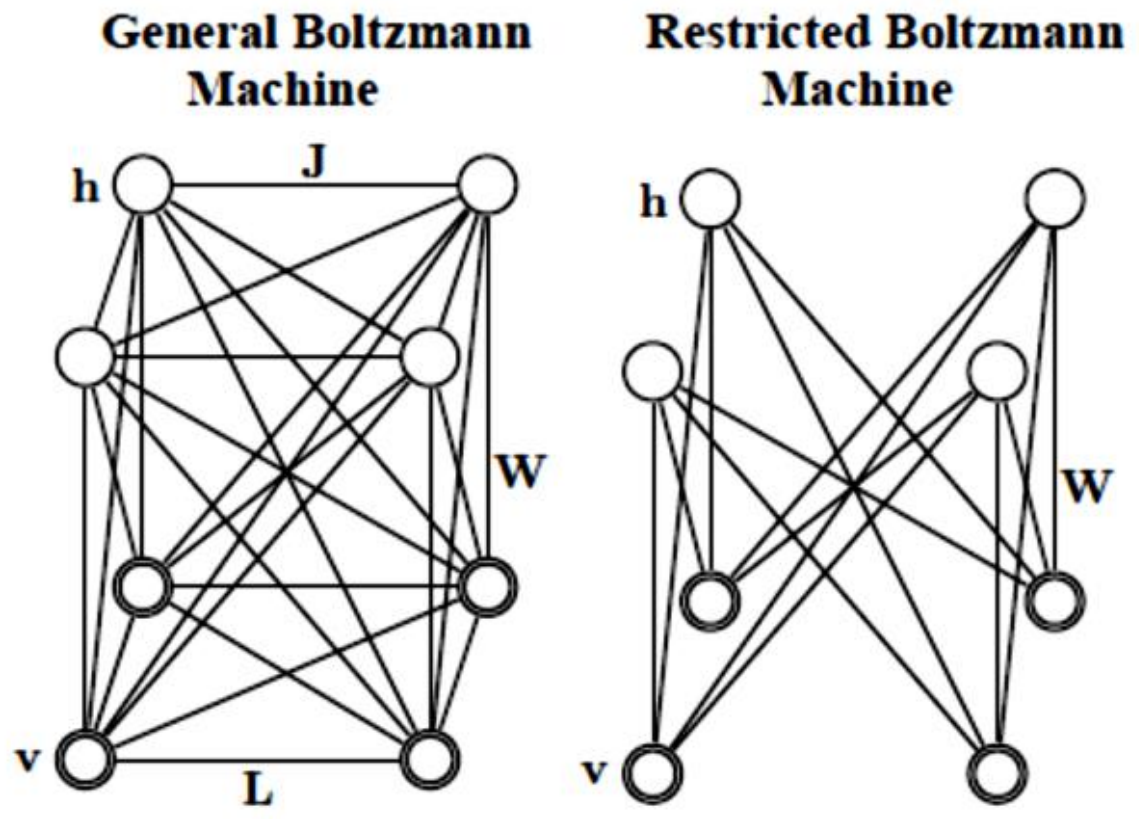

Figure 1. Left hand side figure: a general Boltzmann machine. The top layer shows stochastic binary hidden units and the bottom layer shows stochastic binary visible units. Right hand side

figure: A restricted Boltzmann machine. the joints between hidden units and also between visible units are disconnected [9].

The energy of the joint configuration $\{\mathrm{v}, \mathrm{h}\}$ in restricted Boltzmann machine, with respect to adding bias is given by:

$$E(v,h) = -v^T W h - a^T v - b^T h$$
$$= -\sum_{i=1}^{g_v} \sum_{j=1}^{g_h} W_{ij} v_i h_j - \sum_{i=1}^{g_v} a_i v_i - \sum_{j=1}^{g_h} b_j h_j \qquad (2)$$

Where $W_{ij}$ represents the symmetric interaction term between visible unit $i$ and hidden unit $j$, while $b_i$ and $a_j$ are bias terms for hidden units and visible units respectively. The network assigns a probability value with energy function to each state in visible and hidden units.

Because potential functions in MRFs are strictly positive, it is convenient to express them as exponential and Boltzmann distribution [10]. The joint distribution is defined as the product of potentials, and so the total energy is obtained by adding the energies for potential functions. Therefore joint probability distribution for visible and hidden units can be defined as:

$$P(v,h) = \frac{1}{Z} \exp\big(-E(v,h)\big) \qquad (3)$$

Where $Z$ as partition function or normalization constant, is obtained by summing over all possible pairs of visible and hidden vectors.

$$Z = \sum_v \sum_h \exp\big(-E(v,h)\big) \qquad (4)$$

The probability assigned to a visible vector $v$ by the network, is obtained by marginalizing out hidden vector $h$.

$$P(v) = \sum_h P(v,h) = \frac{1}{Z} \sum_h \exp\big(-E(v,h)\big) \qquad (5)$$

The probability that the network assigns to a training image can be increased by adjusting the weights and biases to lower the energy of that image and to raise the energy of other images, especially those images that have low energies and therefore make a big contribution to the partition function [11]. Therefore, best value for each parameter can be found using the following objective function:

$$maximize_{\{w_{ij},a_i,b_j\}} \frac{1}{m}\sum_{l=1}^{m} \log\left(\sum_{h} P\left(\boldsymbol{v}^{(l)}, \boldsymbol{h}^{(l)}\right)\right) \tag{6}$$

Where the parameter $m$ is the number of training data samples and the aim is to increase the model probability for these training data. Therefore the partial derivative with respect to $w_{ij}$ of the above objective is given by [12] :

$$\frac{\partial}{\partial w_{ij}}\left(\frac{1}{m}\sum_{l=1}^{m} \log\left(\sum_{h} P\left(\boldsymbol{v}^{(l)}, \boldsymbol{h}^{(l)}\right)\right)\right) = \frac{1}{m}\sum_{l=1}^{m}\sum_{h} X_{il} h_j P(h|v = x) - \sum_{v'}\sum_{h'} v'_i h'_j P(v', h') \tag{7}$$

Where $X_{il}$ refers to the $i^{th}$ unit of the $l^{th}$ data instance. The sum on the left hand side can be computed exactly; however the expectation on the right hand side (also called the expectation under the model distribution) is intractable. Therefore other methods are used to estimate this partial derivative. The derivative of the log probability of a training vector with respect to a weight can be computed as follows:

$$-\frac{\partial \log P(v)}{\partial w_{ij}} = < v_i h_j >_{data} - < v_i h_j >_{model} \tag{8}$$

Where the angle brackets are used to denote expectations under the distribution specified by the subscript that follows. This leads to a very simple learning rule for performing stochastic steepest ascent in the log probability of the training data:

$$\Delta w_{ij} = \epsilon \,(< v_i h_j >_{data} - < v_i h_j >_{model}) \tag{9}$$

Where $\epsilon$ parameter is a learning rate. Similarly the learning rule for the bias parameters is:

$$\Delta a_i = \epsilon\,(< v_i >_{data} - < v_i >_{model}) \tag{10}$$

$$\Delta b_j = \epsilon\,(< h_j >_{data} - < h_j >_{model}) \tag{11}$$

Since there are no direct connections between hidden units in an RBM, these hidden units are independent given visible units [11]. This fact is based on MRF properties [10]. Now Given a randomly selected training image $v$, the binary state $h_j$ of each hidden unit $j$, is set to 1 where its probability is:

$$P(h_j = 1|\boldsymbol{v}) = \mathcal{g}\left(b_j + \sum_i v_i w_{ij}\right) \tag{12}$$

Where $\mathcal{g}(x)$ is the logistic sigmoid function $\mathcal{g}(x) = 1/(1 + \exp(-x))$. Therefore $< v_i h_j >_{data}$ can be computed easily.

Since there are no direct connections between visible units in an RBM, it is very easy to obtain an unbiased sample of the state of a visible unit, given a hidden vector

$$P(v_i = 1|\boldsymbol{h}) = \mathcal{g}\left(a_i + \sum_j h_j w_{ij}\right) \tag{13}$$

However computing $< v_i h_j >_{model}$ is so difficult. It can be done by starting from any random state of the visible units and performing sequential Gibbs sampling for a long time. Finally due to impossibility of this method and large run-times, Contrastive Divergence (CD) method is used [13].

RBM has many benefits and has been greatly used in recent years, especially in DBN's. Nowadays many papers wish to improve this model and its performance. In the following section these improvements on computing gradient of log probability of train data are discussed.

## 2.1. Computing gradient of log probability of training data

According to equation (5), the $\log P(v)$ can be expressed as follows [14]:

$$\phi = \log P(v) = \phi^+ - \phi^- \tag{14}$$

$$\phi^+ = \log \sum_h \exp\big(-E(v,h)\big)$$

$$\phi^- = \log Z = \log \sum_v \sum_h \exp\big(-E(v,h)\big)$$

The gradient of $\phi^+$ according to model parameters is a positive gradient and similarly, the gradient of $\phi^-$ according to model parameters is a negative gradient.

$$\frac{\partial \phi^+}{\partial w_{ij}} = v_i . P(h_j = 1|v)$$

$$\frac{\partial \phi^-}{\partial w_{ij}} = P(v_i = 1, h_j = 1) \qquad (15)$$

Computing the positive gradient is simple but computing the negative gradient is intractable and therefore inference methods using sampling are used to compute gradient.

Based on the above sections, the gradient of log probability of training data is obtained from equation (*8*). We must compute $< v_i h_j >_{data}$ and $< v_i h_j >_{model}$ for computing gradient and adjusting parameters according to equation *(9)*. Based on most of the literatures on RBMs, computing $< v_i h_j >_{data}$ is called positive phase, and computing $< v_i h_j >_{model}$ is called negative phase corresponding to positive gradient and negative gradient respectively.

Since there is no interconnections between hidden units and they are independent, $< v_i h_j >_{data}$ can easily be computed by considering the visible units $v$ (that their values have been determined by training data) and assigning the value 1 to each hidden unit with the probability of $P\big(h_j = 1\big|v\big)$ regarding to equation *(12)*.

The main problem resides in the negative phase. In practice, the difference between different DBN learning methods (e.g. Contrastive Divergence or Persistent Contrastive Divergence) is in sampling in their negative phase [15].

To compute $< v_i h_j >_{model}$, Gibbs sampling method may be used. This method starts with random values in visible units and Gibbs sampling steps should continue for a long time. Each Gibbs sampling step leads to updating of all hidden units according to equation (*12*) and then updating all visible units according to equation (*13*) (see Figure 2). Indeed, Gibbs sampling is a method for obtaining a good sample from joint distribution on $v$ and $h$ in this model.

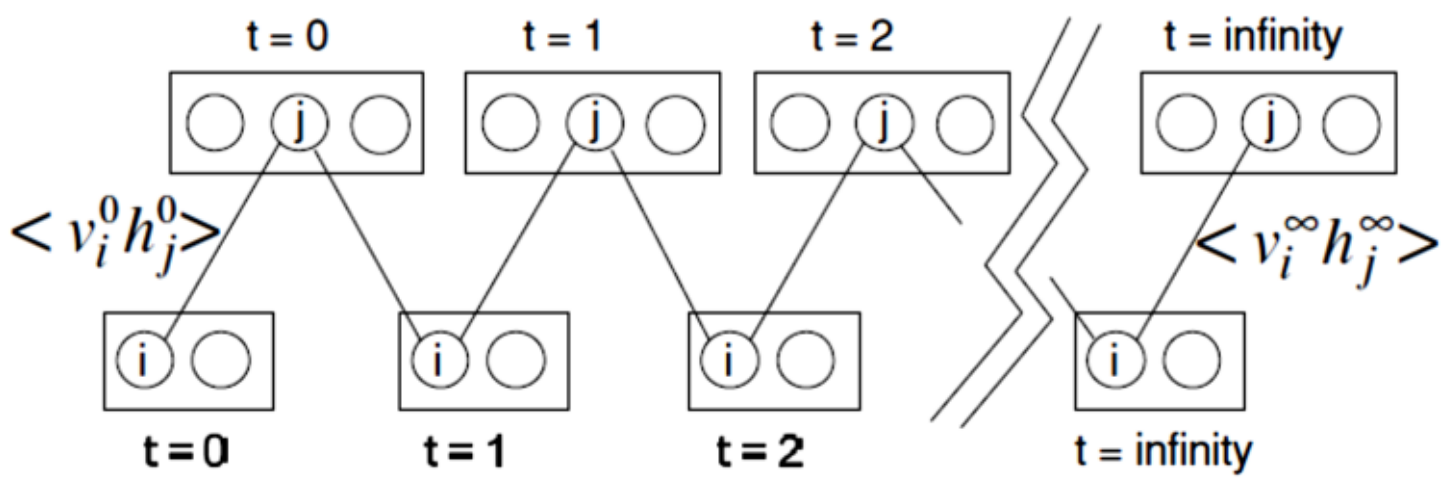

Figure 2: Gibbs sampling. Each Gibbs sampling step means updating of all hidden units according to equation (12) and then updating all visible units according to equation (13). The chain is initialized by setting the binary states of the visible units to be the same as a data vector [4].

### 2.1.1. Contrastive Divergence (CD)

Since Gibbs sampling method is slow, Contrastive Divergence (CD) algorithm is used [13]. In this method visible units are initialized using training data. Then binary hidden units are computed according to equation (*12*). After determining binary hidden unit states, $v_i$ values are recomputed according to equation (*13*). Finally, probability of hidden unit activations is computed and using these values of hidden units and visible units, $< v_i h_j >_{model}$ is computed. The computation steps in $CD_1$ method is graphically illustrated in Figure 3.

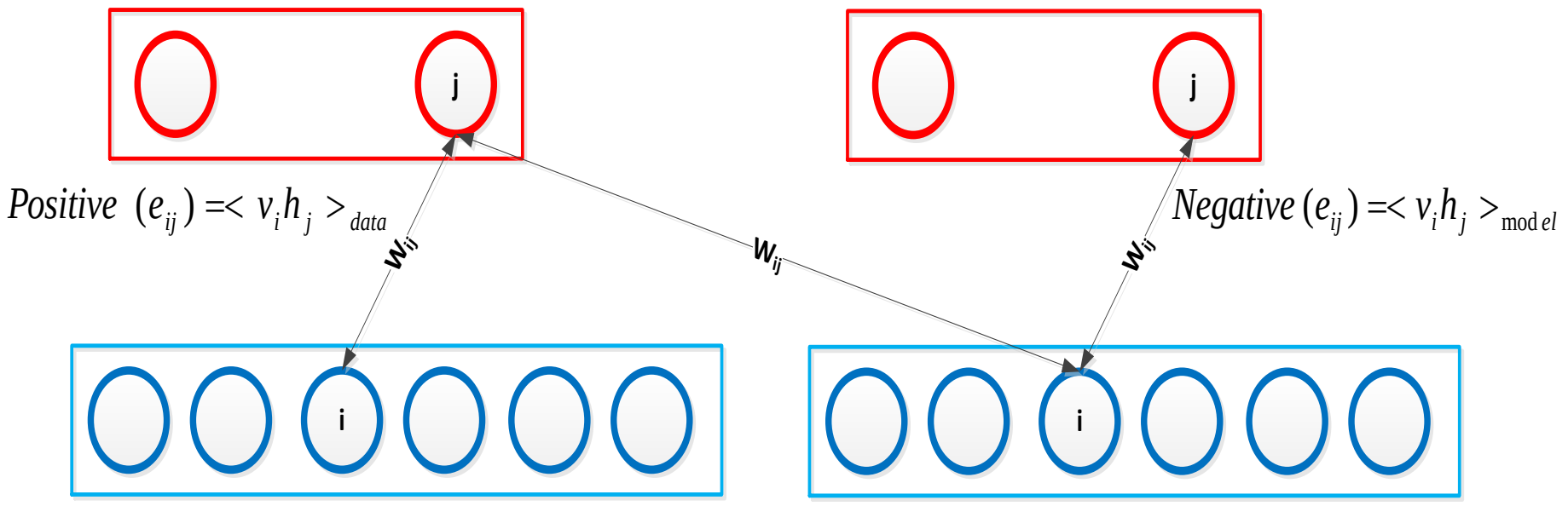

Figure 3: Computation steps in $CD_1$ method.$Positive\ (e_{ij})$ is related to computing $< v_i h_j >_{data}$ for $e_{ij}$ connection.

Although $CD_1$ method is not a perfect gradient computation method, but its results are acceptable [13]. By repeating Gibbs sampling steps, $CD_k$ method is achieved. The k parameter is the number of repetitions of Gibbs sampling steps. This method has a higher performance and can compute gradient more exactly [16].

### 2.1.2. Persistent Contrastive Divergence (PCD)

Whereas $CD_k$ has some disadvantages and is not exact, other methods are proposed in RBM. One of these methods is PCD that is very popular [17]. Unlike CD method that uses training data as initial value for visible units, PCD method uses last chain state in the last update step. In other words, PCD uses successive Gibbs sampling runs to estimate $< v_i h_j >_{model}$. Although all model parameters are changed in each step, but can receive good samples from model distribution with a few Gibbs sampling steps because the model parameters change slightly [18]. Many persistent chains can be run in parallel and we will refer to the current state in each of these chains as new sample or a “fantasy” particle [9], [17].

### 2.1.3. FEPCD (Free Energy in Persistent Contrastive Divergence)

Since in an RBM each unit in a layer is independent from other units in other layers, therefore Gibbs sampling is a proper method. But in order to obtain appropriate samples from the model, Gibbs sampling needs to be run for many times and this is impossible. Therefore different methods as CD or PCD have been proposed. In another paper by authors, a new method for generating better samples as described later has been proposed [19].

In PCD method, as described before, many persistent chains can be run in parallel and we will refer to the current state in each of these chains as a “fantasy” particle. Chain selection in this method is blind and the best one may not be selected. If we can define a criterion for goodness of a chain, samples and therefore computing gradient will be more accurate.

The proposed criterion for selecting the best chain is the free energy of visible sample $v$ which is defined as follows [11]:

$$P(v) = \frac{1}{Z} e^{-F(v)} = \frac{1}{Z} \sum_{h} e^{-E(v,h)} \tag{16}$$

where $F(v)$ is free energy. Therefore $F(v)$ can be computed as follows [11]:

$$F(v) = -\sum_{i} v_i a_i - \sum_{j} q_j I_j + \sum_{j} \left( q_j \log q_j + (1 - q_j) \log(1 - q_j) \right) \tag{17}$$

Where $I_j = b_j + \sum_i v_i w_{ij}$ is equal to sum of inputs to hidden unit $j$ and $q_j = g(I_j)$ is equal to activation probability of hidden unit $h_j$ given $v$ and $g$ is logistic function. An equivalent and simpler equation for computing $F(v)$ is as follows:

$$F(v) = -\sum_i v_i a_i - \sum_j \log(1 + e^{I_j}) \qquad (18)$$

### 2.2. Discriminative RBM

An RBM can also model the joint distribution of the inputs and associated target classes. In this toolbox, we use this joint model, which is depicted in the following figure [20].

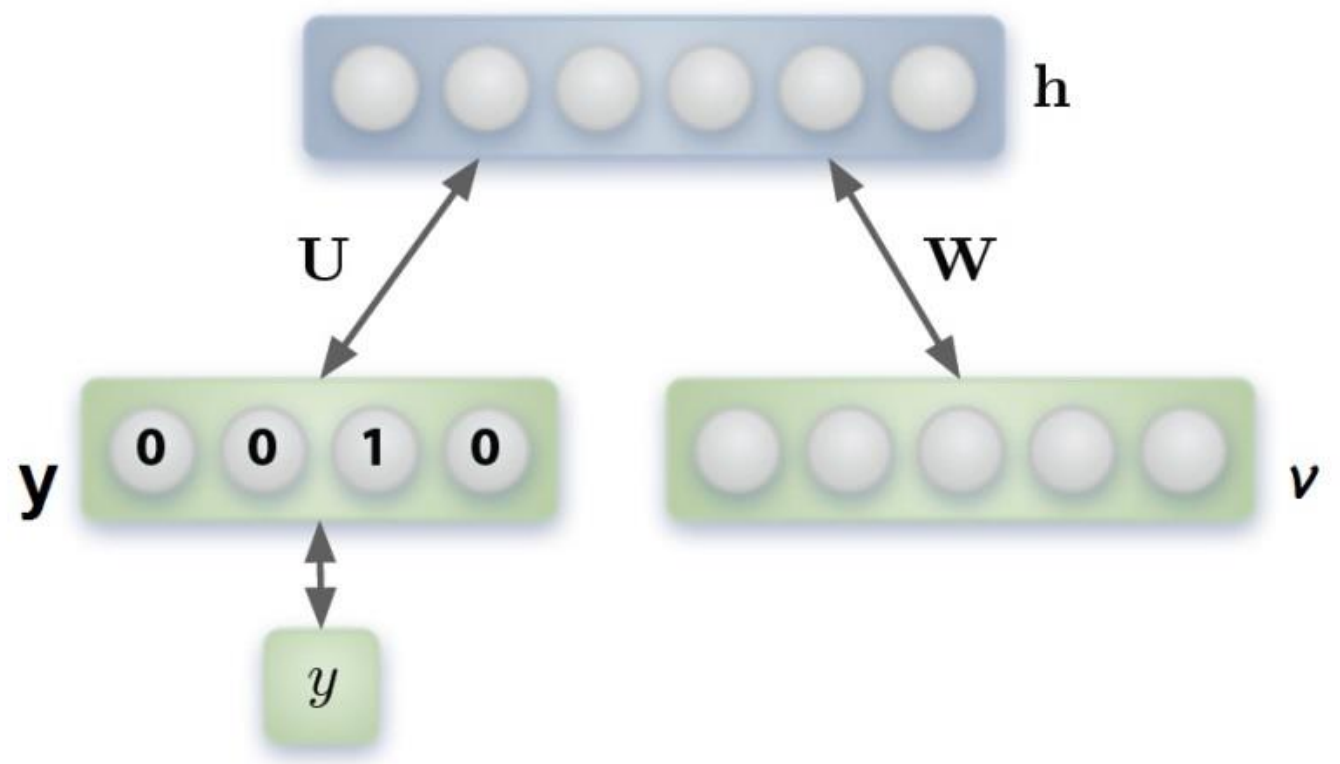

Figure 4: Restricted Boltzmann Machine modeling the joint distribution of inputs $z$ and target class $y$. The current state of the hidden units is labeled by $h$ [20].

This method aims to train a joint density model using a single RBM that has two sets of visible units. In addition to the units that represent a data vector, there is a “softmax” label unit that represents the class. After training, each possible label is tried in turn with a test vector and the label that gives lowest free energy is chosen as the most likely class [11].

### 2.3. Deep Belief Network

After an RBM has been learned, the activity values of its hidden units (when they are being driven by data) can be used as the ‘training data’ for learning a higher-level RBM [21]. The idea behind DBN is to allow each RBM model in the sequence to receive a different representation of the data. According to Figure 5 the model performs a nonlinear transformation on its input vectors and produces as output, the vectors that will be used as input for the next model in the sequence [4].

After layer-by-layer pre-training in DBN, we use back-propagation technique through the whole classifier to fine-tune the weights for optimal classification.

Pretraining helps generalization and the very limited information in the data is used only to slightly adjust the weights found by pretraining [3].

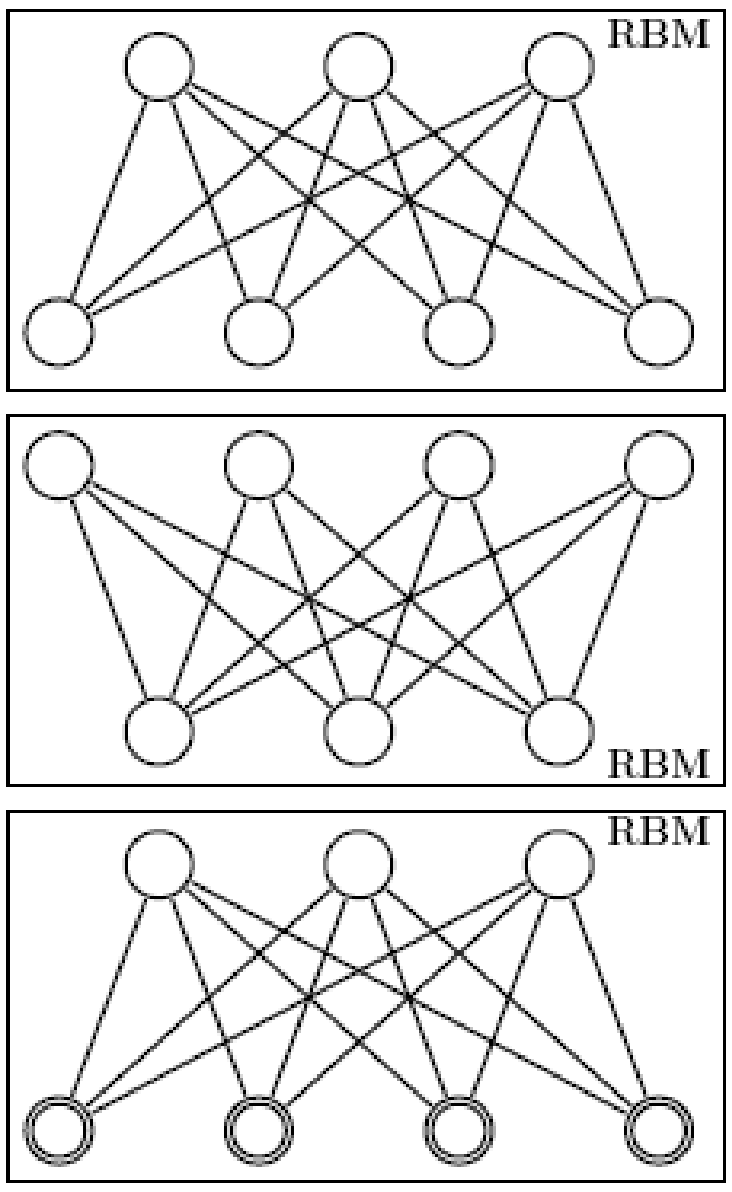

Figure 5: A DBN model. Each RBM model performs a nonlinear transformation on its input vectors and produces as output, the vectors that will be used as input for the next RBM model in the sequence [22].

# 3. An object oriented toolbox for Deep Belief Networks(*DeeBNet*)

The *DeeBNet*[9] is an object oriented MATLAB toolbox to provide tools for conducting research using Deep Belief Networks. The toolbox has two packages with some classes and functions for managing data and sampling methods and also has some classes to define different RBMs and DBN. The following sections describe these packages and classes in more details. The Figure 6 shows relationships between implemented classes.

---

[9] **Dee**p **B**elief **Net**work

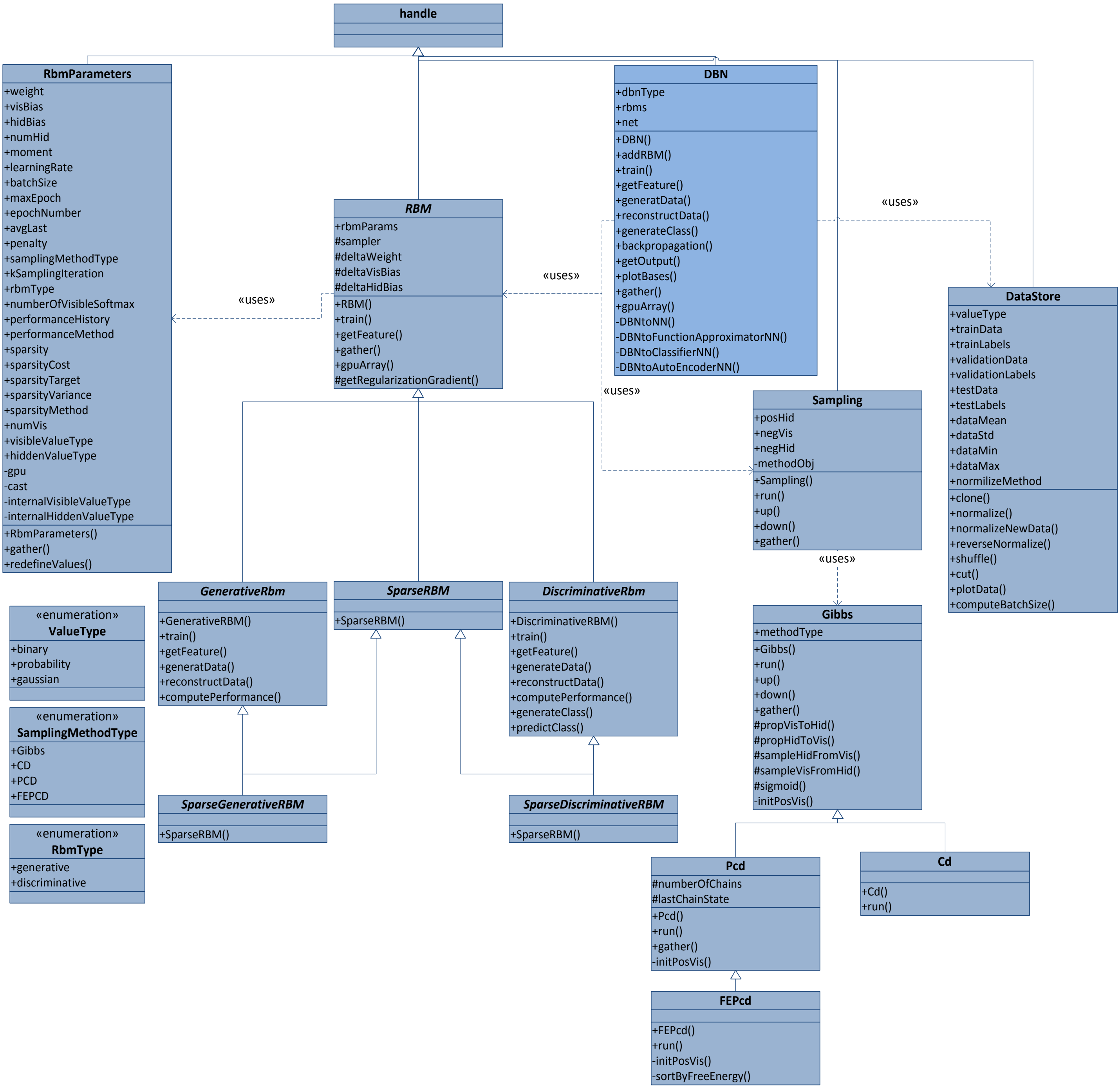

Figure 6: Relationships between implemented classes in *DeeBNet* toolbox

## 3.1. Basic classes

In this section, the basic classes are defined. These classes will be used in RBM and DBN. The first class is ValueType that is an enumeration. This class define different types of units in DBN. These defined types can be binary (with 0 or 1 value), probability (with values in interval [0,1]) and Gaussian (with any real values with zero mean and unit variance).

*RbmType* is also an enumeration. This class defines different types of RBMs. These defined types are generative (use data without their labels) and discriminative (need data with their labels and can classify data).

Another important class is *RbmParameters* that includes all parameters of an RBM such as weight matrix, biases, learning rate, etc. Most of these parameters are defined in [11].

*DataClasses* package has one class to manage train, test and validation data. The *DataStore* class has some useful functions such as *normalize* and *shuffle* function for normalizing and shuffling data. Also it provides the *cut* function to cut training data and choose a part of it as training data. Finally the *plotData* function can be used for plotting some parts of data. It is useful for compare data before and after some processing stages (see Figure 7).

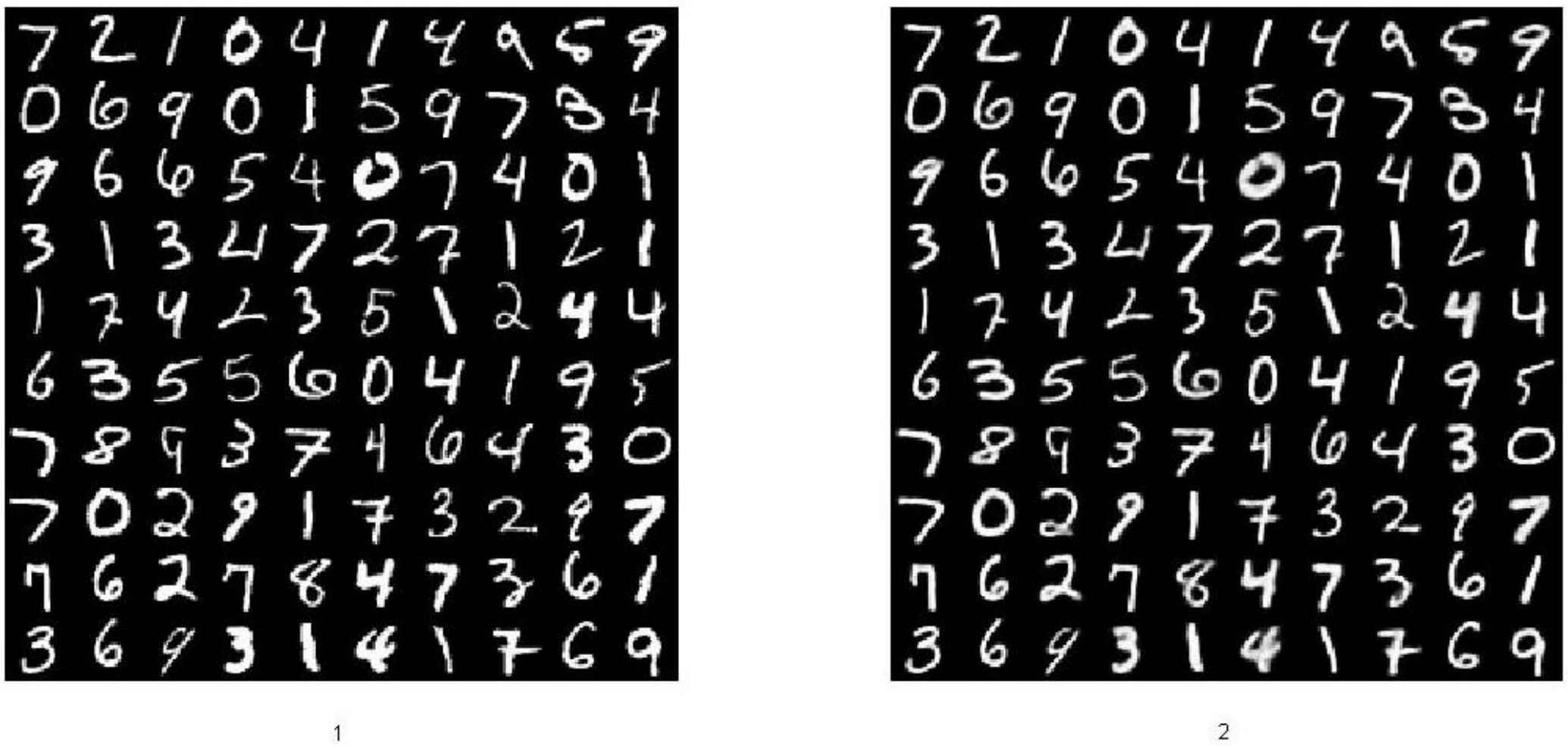

Figure 7: Plotting 100 samples with *plotData* function in *DataStore* class. The first image is 100 samples from MNIST dataset and the second one is reconstructed samples with a DBN model. The related code is in "*test_plotData.m*" file.

The second package includes the implementation of some different sampling methods. These sampling methods are *Gibbs*, *CD*, *PCD* and *FEPCD*. In *Gibbs* class we can generate samples from an RBM model with random initialization samples (see section 2.1). Also this class is a parent class for other sampling classes. In the *CD* (Contrastive Divergence) class, we can generate samples from an RBM model with training samples initialization (see section 2.1.1). This class inherits from *Gibbs* class.

In *PCD* (Persistent Contrastive Divergence) class, samples can generated from an RBM model. Unlike *CD* method that uses training data as initial value for visible units, *PCD* method uses last chain state in the last update step. Also this class inherits from Gibbs class (see section 2.1.2). In this class many persistent chains can be run in parallel and we will refer to the current state in each of these chains as new sample or a “fantasy” particle.

In *FEPCD* (Free Energy in Persistent Contrastive Divergence) class, we define a criterion for goodness of a chain and therefore generated samples and gradient computation will be more accurate. The proposed criterion for selecting the best chain is the free energy of visible sample $v$ (see section 2.1.3). This class inherits from *PCD* class.

Finally the *Sampling* class is an interface class for using implemented sampling classes. Other classes can use implemented sampling classes such as CD or PCD with this useful class. In this class we use *SamplingMethodType* class that is an enumeration and contains types of sampling methods that are used in RBM.

## 3.2. RBM classes

The toolbox has six types of RBM classes. The first one, *RBM* class, is an abstract class that defines all necessary functions (such as training method) and features (like sampler object) in all types of RBMs and therefore we can't create an object from it. Other RBM classes are inherited from this abstract class.

The second one is *GenerativeRBM* class. This class has been used as a generative model and can model many different types of data. Their most important use is as learning modules that are composed to form DBNs (see section 02). The *GenerativeRBM* class has many methods like *train*, *getFeature*, *generateData*, *reconstructData*, etc. The ***train*** method takes a *DataStore* object (that has training, validation and test data) and modifies the RBM parameters. The termination condition is the number of training epochs. The *getFeature* method, extracts features (or activity in hidden layer) from data. In other words this method samples hidden units from visible units with determined sampling method.

The *generateData* method can generate values of visible units from determined hidden values (or extracted features). Similar to *getFeature* method, *generateData* samples visible units from hidden units with determined sampling method. Figure 8 shows some outputs of the method. These results have been obtained from an RBM with 250 hidden units that has been trained on MNIST dataset. In this experiment, after extracting 250 feature from 9 MNIST images (28*28 pixel), the new images have been generated from extracted features. According to Figure 8, by increasing $k$ (number of sampling iterations), the generated images will be more natural and more similar to data distribution.

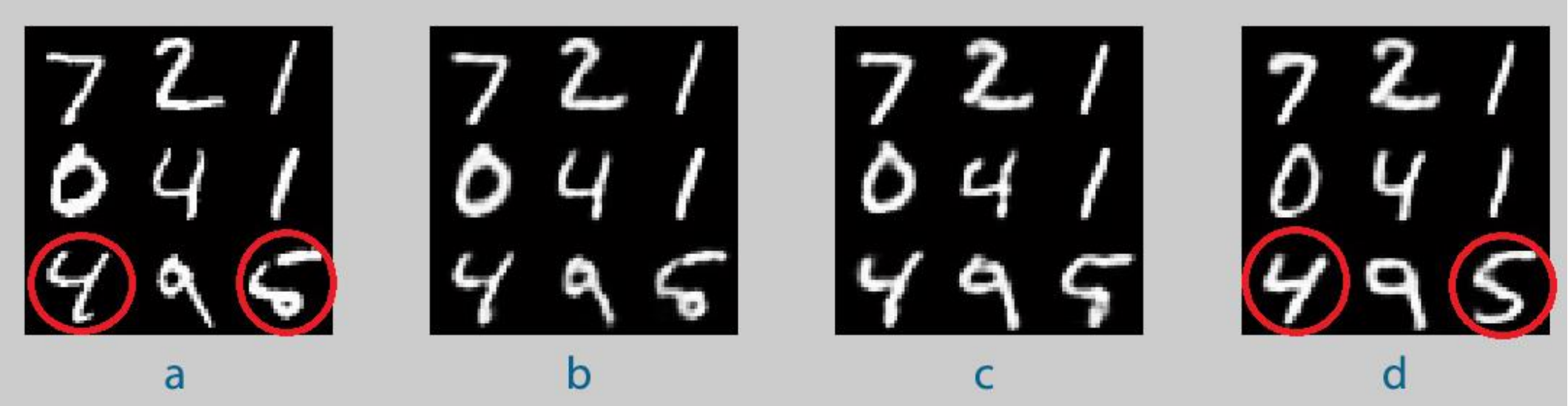

Figure 8: Results from an RBM with 250 hidden units that has been trained on MNIST dataset. In this experiment, after extracting 250 features from 9 MNIST images (28*28 pixel), the new images have been generated from extracted features. (a) 9 MNIST images. (b) Generated images from extracted features with $k = 1$ sampling iteration. (c) Generated images from extracted features with $k = 10$ sampling iterations. (d) Generated images from extracted features with $k = 100$ sampling iterations. The related code is in "*test_ generateData.m*".

The last useful method is *reconstructData*. This method is used for reconstructing input data. Indeed the method reconstruct data by extracting features from input data and then generating data from extracted features. In Figure 9 this method has been used to reduce noise in images. According to Figure 9, Gaussian noise has been reduced after reconstructing corrupted images.

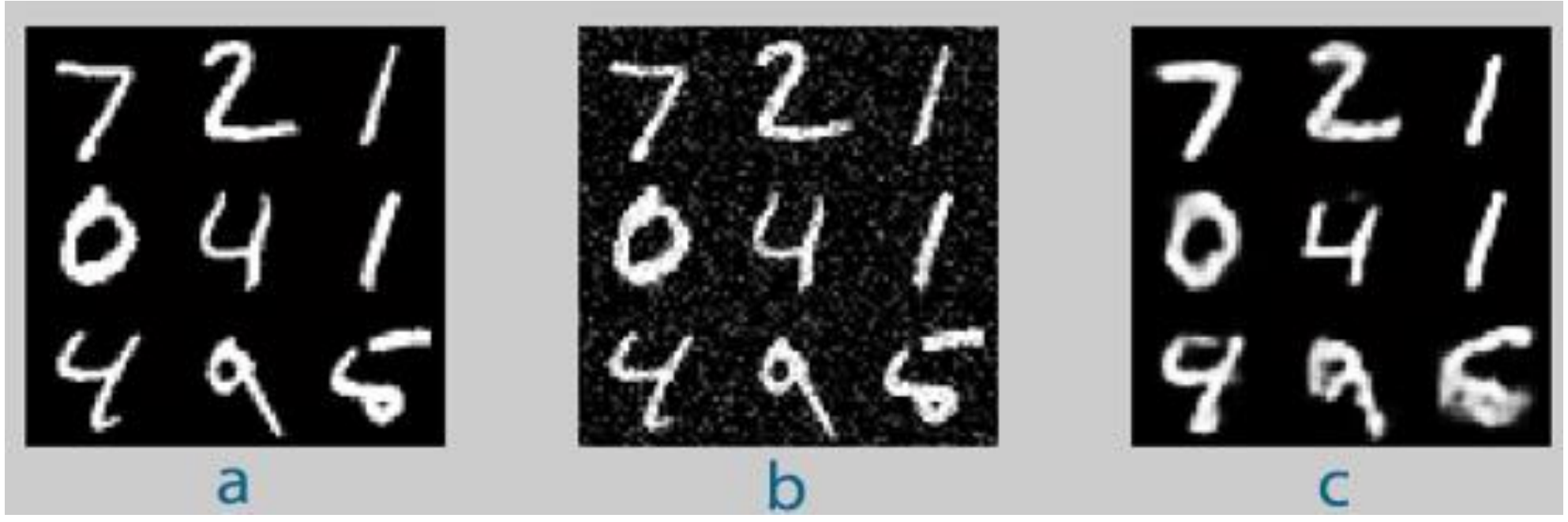

Figure 9: reducing noise from corrupted images using *reconstructData* method. (a) 9 MNIST images. (b) Corrupted data with Gaussian noise with zero mean and 0.02 variance. (c) Reconstructed images from corrupted images. Gaussian noise has been reduced after reconstructing corrupted images. The related code is in "*test_reconstructData2.m*".

The third RBM class is *DiscriminativeRBM*. With some changes, we can convert generative RBM to a discriminative RBM that can classify data (see section 2.2). This class includes methods like methods in *GenerativeRBM* class. Two different methods are *generateClass* and *predictClass*. The *generateClass* can generate data with a specified class number (or label). According to Figure 10, the model can generate different images with only activating label unit in model. Note that the model can't generate images for two digits (2 and 8) using only activating label unit.

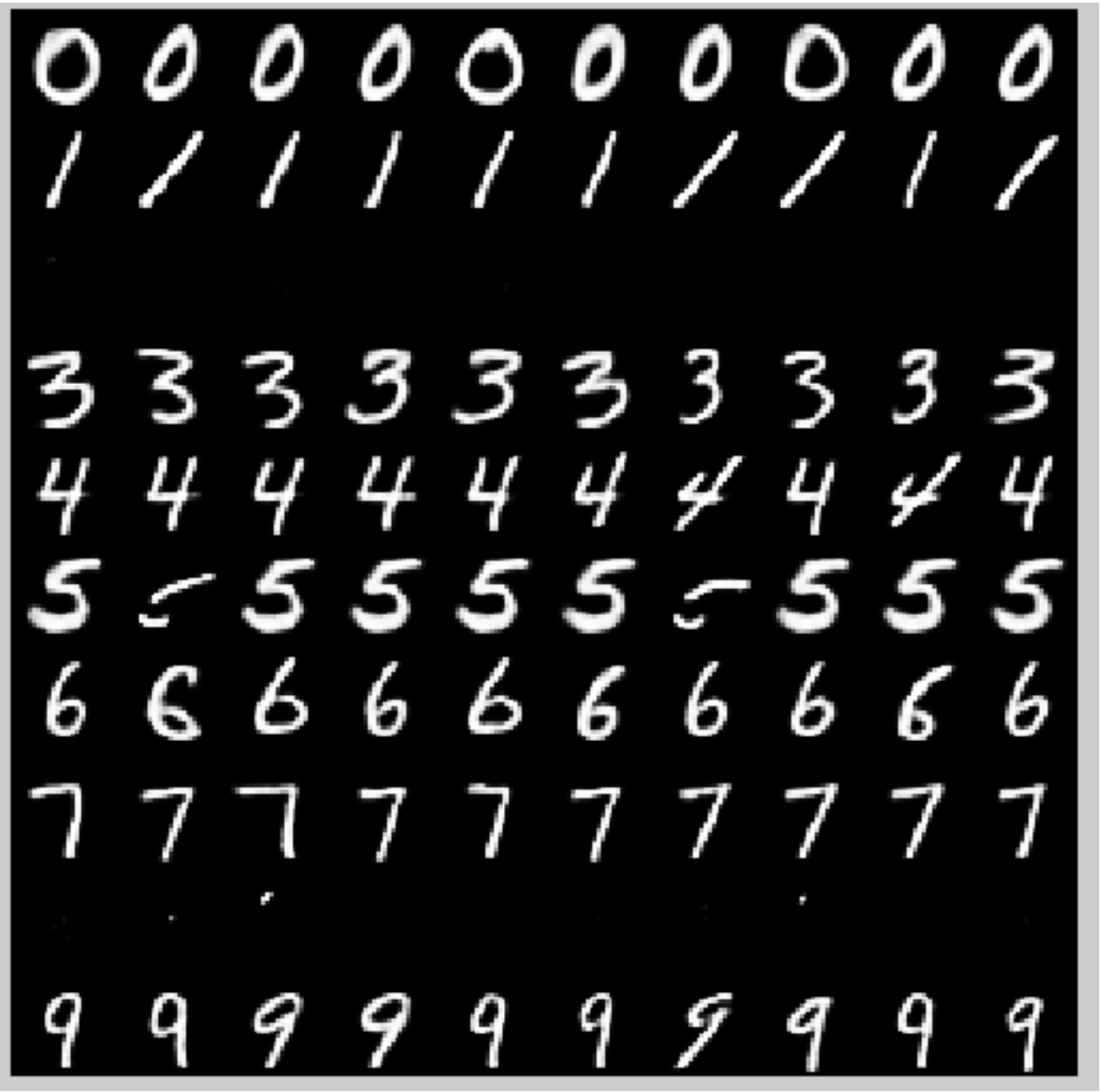

Figure 10 : synthesized images with *generateClass* method. Generating different images with only activating label unit in model. Using this method, the model can generate different images by only activating label unit in model. Note that the model can't generate images for two digits (2 and 8) by only activating label unit. The related code is in "*test_generateClass.m*".

The other different method is *predictClass*. This method can predict class number (or label) of input data. The first method (*byFreeEnergy*) is to train a joint density model using a single RBM that has two sets of visible units. In addition to the units that represent a data vector, there is a "softmax" label unit that represents the class. After training, each possible label is tried in turn with a test vector and the one that gives lowest free energy is chosen as the most likely class [11]. The second method (*bySampling*) is to reconstruct data and return most activated softmax unit (that correspond to a label). Usually the *byFreeEnergy* is more accurate but is more time consuming.

Another three RBM classes are *SparseRBM*, *SparseGenerativeRBM* and *SparseDiscriminativeRBM*. The first one, *SparseRBM* class, is an abstract class that define gradient of regularization term for different sparsity methods such as quadratic sparse RBM, rate distortion sparse RBM and normal sparse RBM [23].

The *SparseGenerativeRBM* and *SparseDiscriminativeRBM* classes combine generative RBM or discriminative RBM with sparse RBM features in a separate class that *GenerativeRBM* and *DiscriminativeRBM* can be sparse.

### 3.3. DBN class

DBN is a generative model that is composed of multiple layers of RBMs (see section 2.3). The class architecture allows using different RBM classes to create an arbitrary DBN and utilizes back-propagation after DBN training if needed. A DBN can be used as an *autoEncoder* or *classifier*.

An *autoEncoder* DBN may be used to create a generative model and can be used in many applications such as feature extraction. Figure 11 shows an *autoEncoder* DBN with two RBM layers. The last layer hidden units can be used as a feature vector that has been extracted from input visible data.

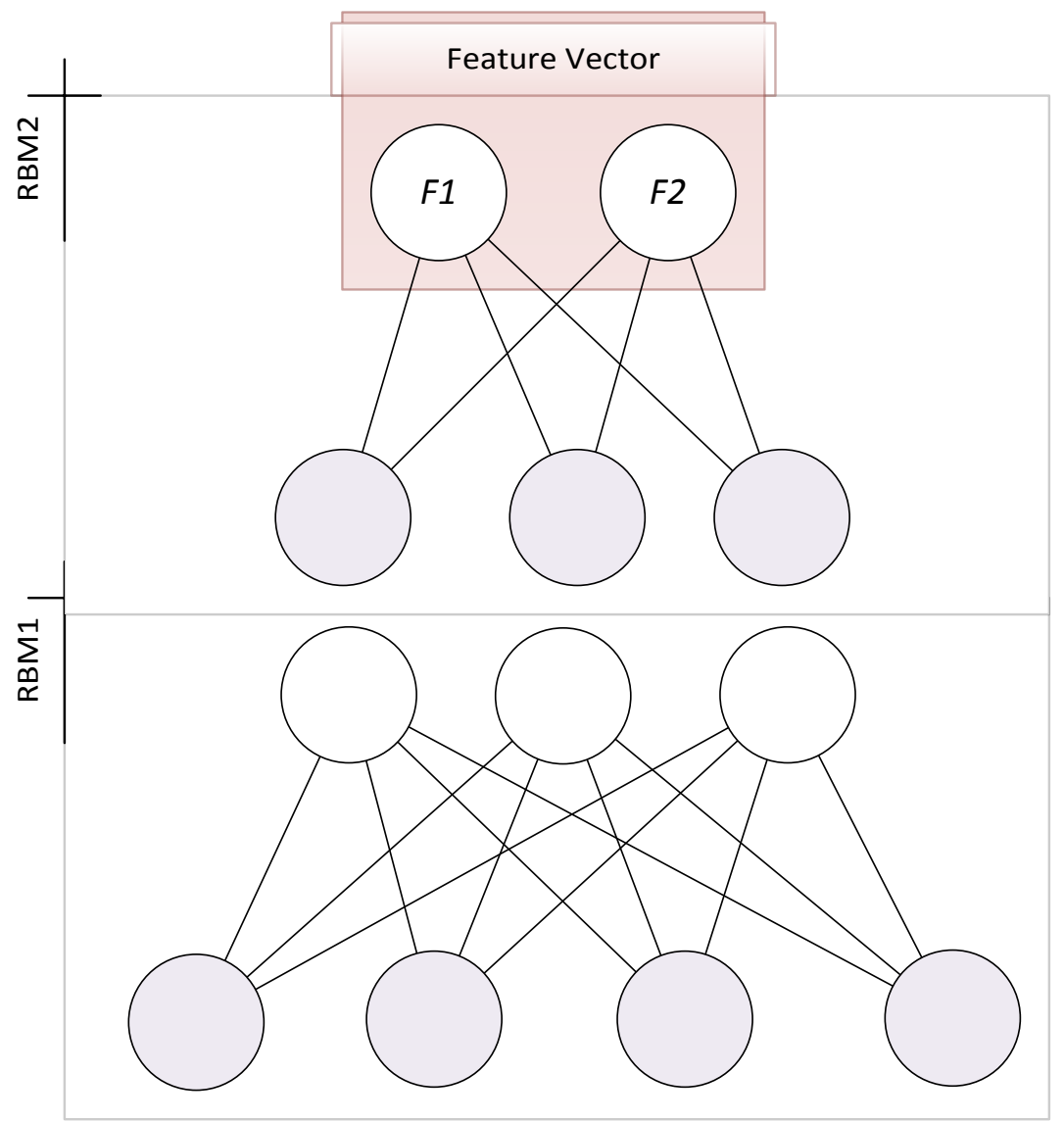

Figure 11: An *autoEncoder* DBN with two RBM layers.

A DBN can also be used as a *classifier*. The goal of *classifier* DBN is to obtain labels from input data. In this type of DBN, we need a discriminative RBM in last layer as a classifier RBM (see section 2.2). Figure 12 shows a *classifier* DBN with two RBM layers where the last RBM is a discriminative RBM.

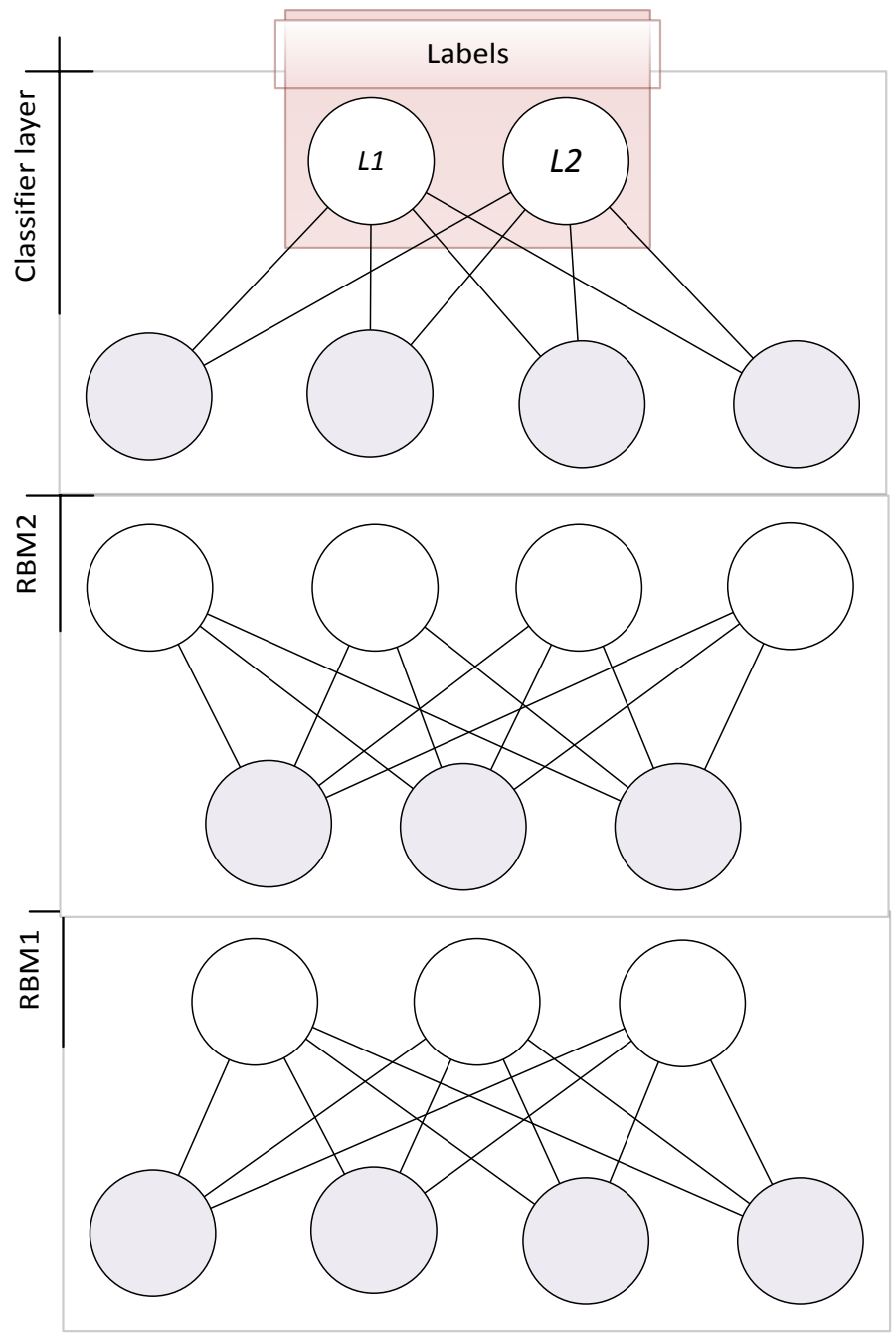

Figure 12: A *classifier* DBN with two RBM layers. The last RBM is a discriminative RBM.

The DBN class has some useful methods like *addRBM*, *train*, *getFeature*, *backpropagation*, *getOutput*, *plotBases*, etc. The *addRBM* method is used to stack RBMs. This method add each defined RBM (with *RbmParameters* object) to its DBN.

The *train* method trains DBN, layer by layer. In other words, this method trains RBMs one after another and uses their extracted features for training in the next RBM.

The "*getFeature"* method is used to extract features from input data. This method extracts features layer by layer and returns hidden units activation values in last hidden layer as extracted feature (see Figure 11).

Figure 13 shows extracted features in a DBN on MNIST dataset. The features were produced by a 784-1000-500-250-3 *autoEncoder* DBN that maps input images (784 pixel) to 3 features.

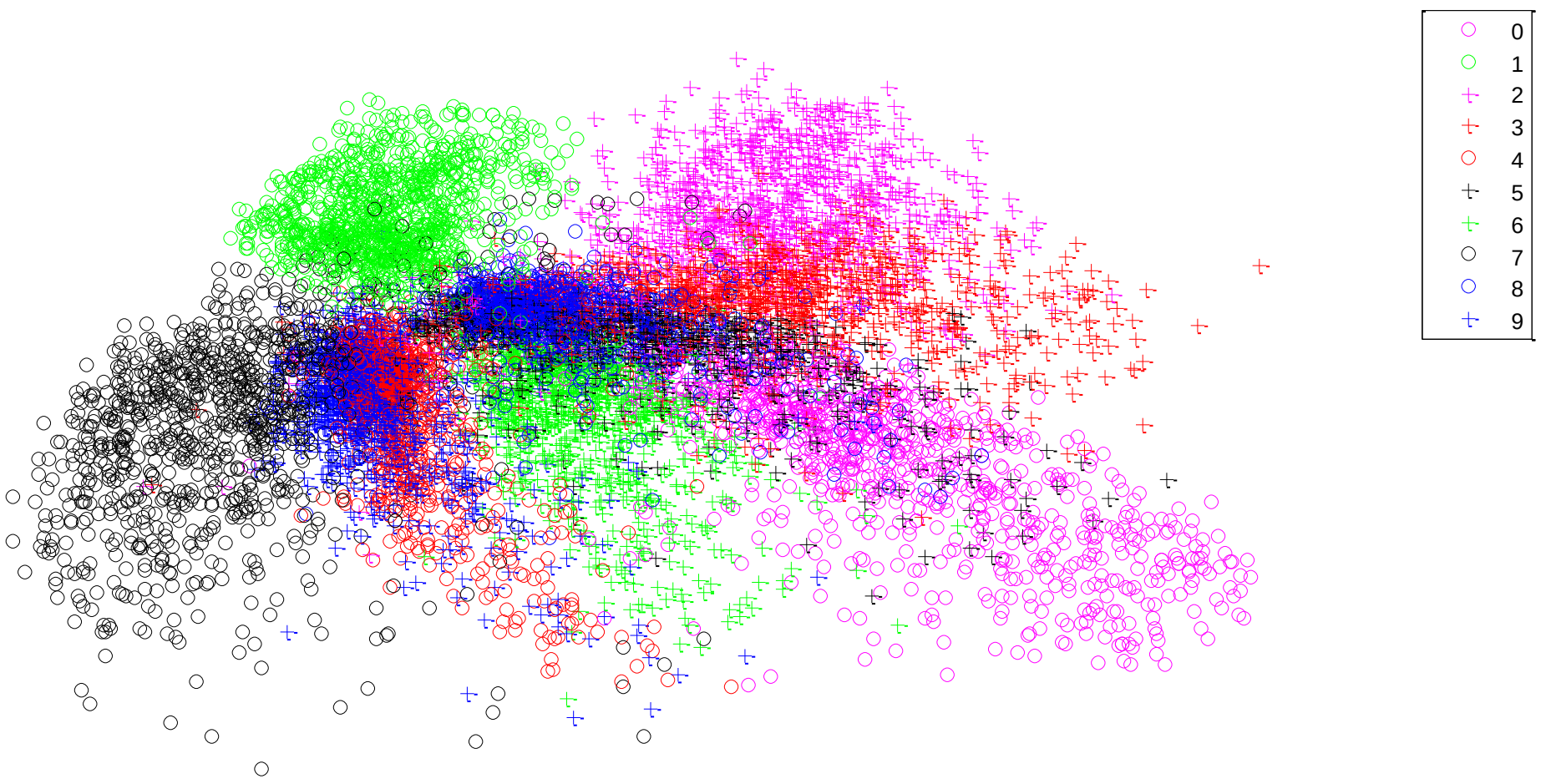

Figure 13: extracted features in a DBN on MNIST dataset. The features produced by a 784-1000-500-250-3 *autoEncoder* DBN that maps input images (784 pixel) to 3 features. The related code can be found in "*test_getFeatureMNIST.m*"file.

In another test, ISOLET dataset is used [24]. In ISOLET data set, 150 subjects utter twice the name of each letter of the alphabet. There are 7797 examples in total, referred to as isolet1-isolet5 (6238 training examples and 1559 test examples). Figure 14 shows extracted features in a DBN on ISOLET dataset. The features produced by a 617-2000-1000-500-250-2 and a 617-2000-1000-500-250-3 *autoEncoder* DBN that maps input data (617 features) to 2 or 3 features.

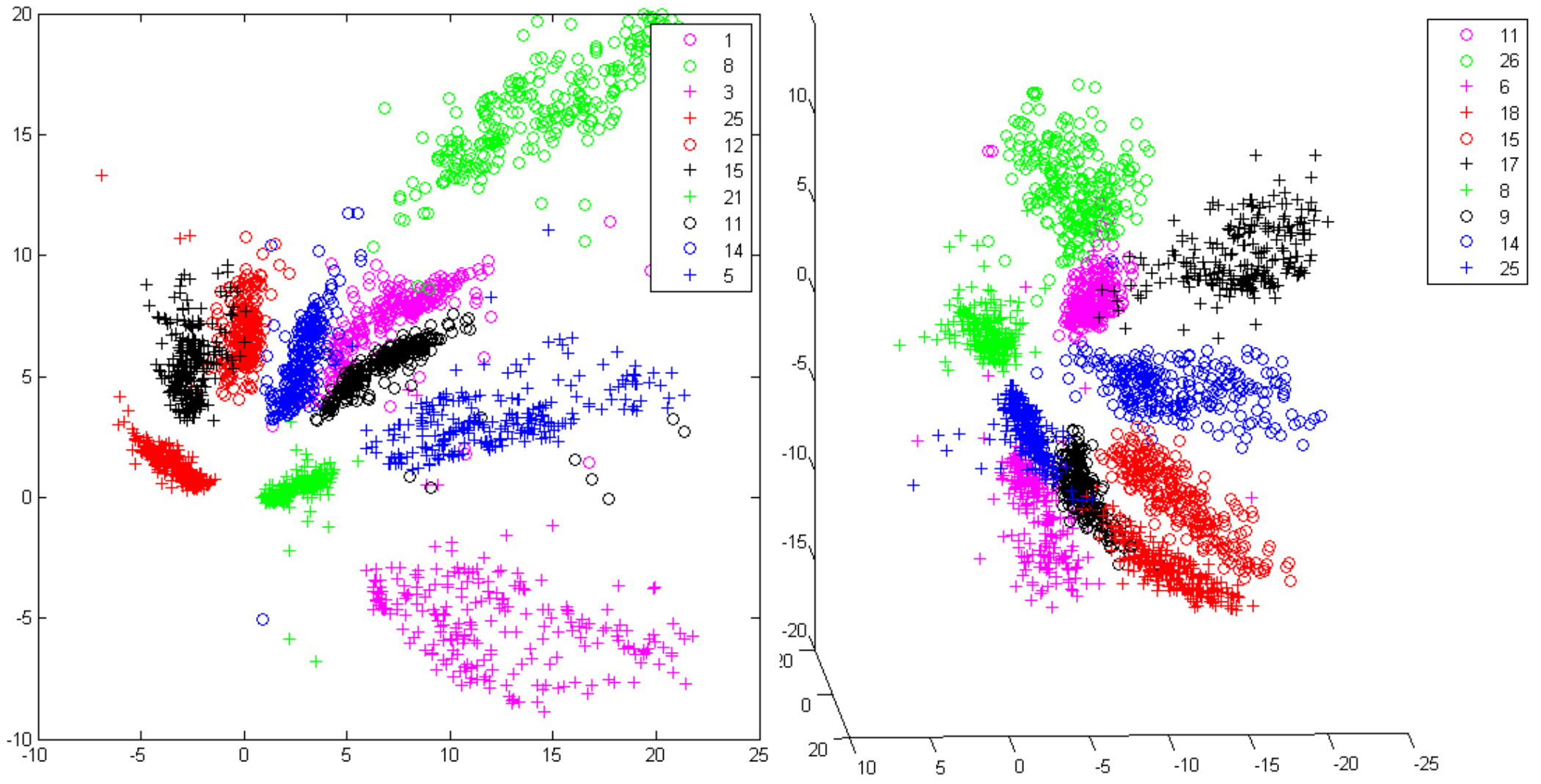

Figure 14: extracted features from a DBN on ISOLET dataset with 617 features and 26 different classes (26 different spoken letters). Ten randomly selected letters are shown. Left: The features produced by a 617-2000-1000-500-250-2 *autoEncoder* DBN. Right: The features produced by a 617-2000-1000-500-250-3 *autoEncoder* DBN. The related code is in "*test_getFeatureISOLET.m*".

Also in another test, 20 Newsgroups dataset is used. The 20 Newsgroups[10] dataset is organized into 20 different newsgroups, each corresponding to a different topic. The 20 Newsgroups dataset has become a popular data set for experiments in text applications of machine learning techniques, such as text classification and text clustering. Figure 15 shows extracted features in a DBN on 20 Newsgroups dataset. The features produced by a 5000-500-500-250-3 *autoEncoder* DBN that maps input data (5000 features) to 3 features.

According to Figure 13, Figure 14 and Figure 15, DBN can obtain good features with acceptable discrimination between them. Note that these features has been learnt without using their labels.

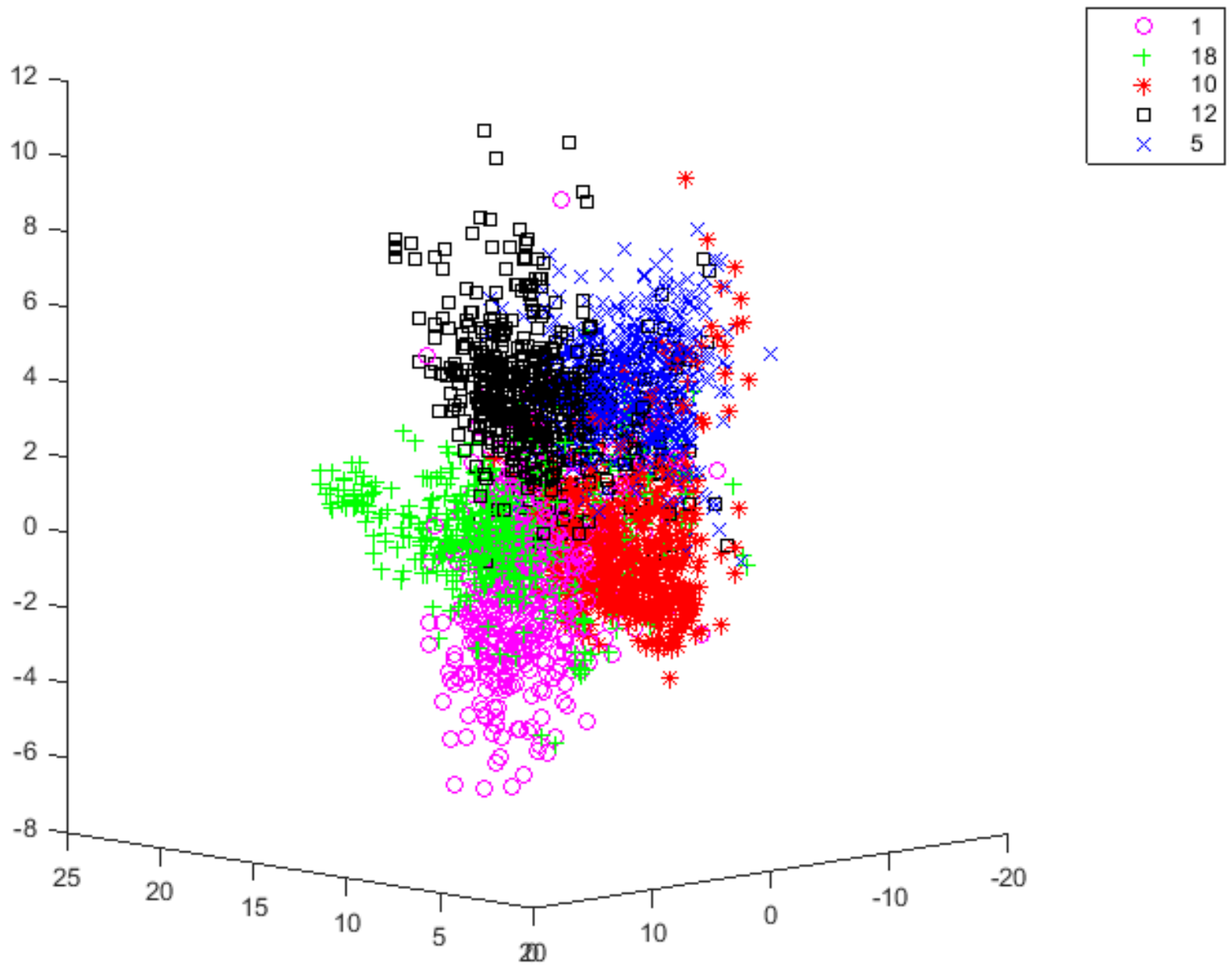

Figure 15: extracted features from a DBN on 20 Newsgroups dataset with 5000 features and 20 different classes (20 different newsgroups). In this figure, five selected newsgroups are shown. The features produced by a 5000-500-500-250-3 *autoEncoder* DBN. The related code is in "*test_getFeature20Newsgroups.m*".

The next useful method is *backpropagation* method. This method uses back-propagation algorithm to fine-tune pertained parameters. Our toolbox uses MATLAB neural network toolbox. Hence the method first converts a DBN to a MATLAB neural network object (according to DBN type) and then uses its back-propagation algorithm.

---

[10] Available online at "http://qwone.com/~jason/20Newsgroups"

Figure 16 shows, how a DBN with a discriminative RBM in last layer converts to a MATLAB neural network structure. In this conversion, the softmax units in discriminative RBM and their corresponding weights are set as output neural network layer.

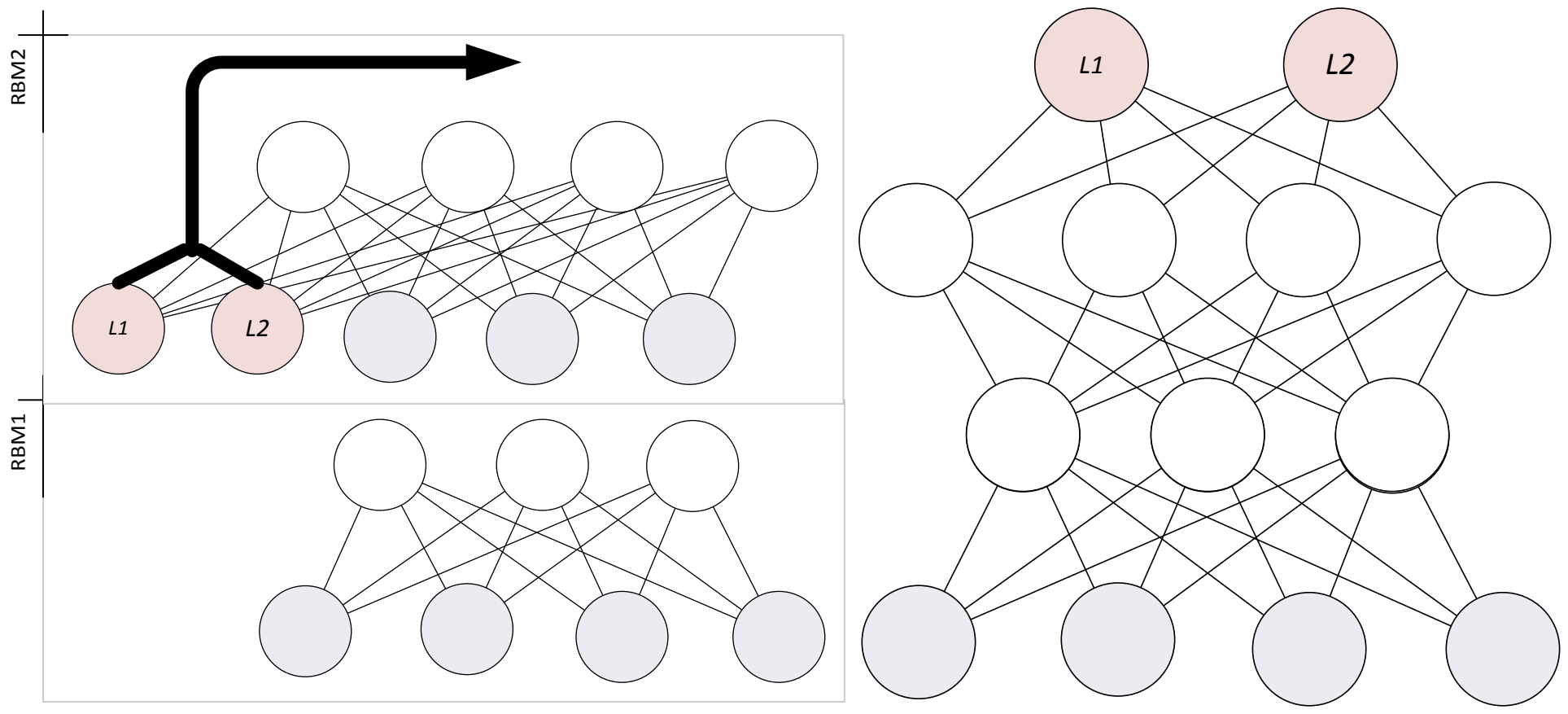

Figure 16: Conversion of a *classifier* DBN to a MATLAB neural network structure. Left: A DBN with a discriminative RBM in last layer. Right: A neural network structure with softmax units and their weights in DBN as output layer.

In an *autoEncoder* DBN, conversion to neural network structure is done differently. Figure 17 shows, how we add an upside down DBN to reconstruct input data [3]. This neural network structure can be fine-tuned using back-propagation algorithm.

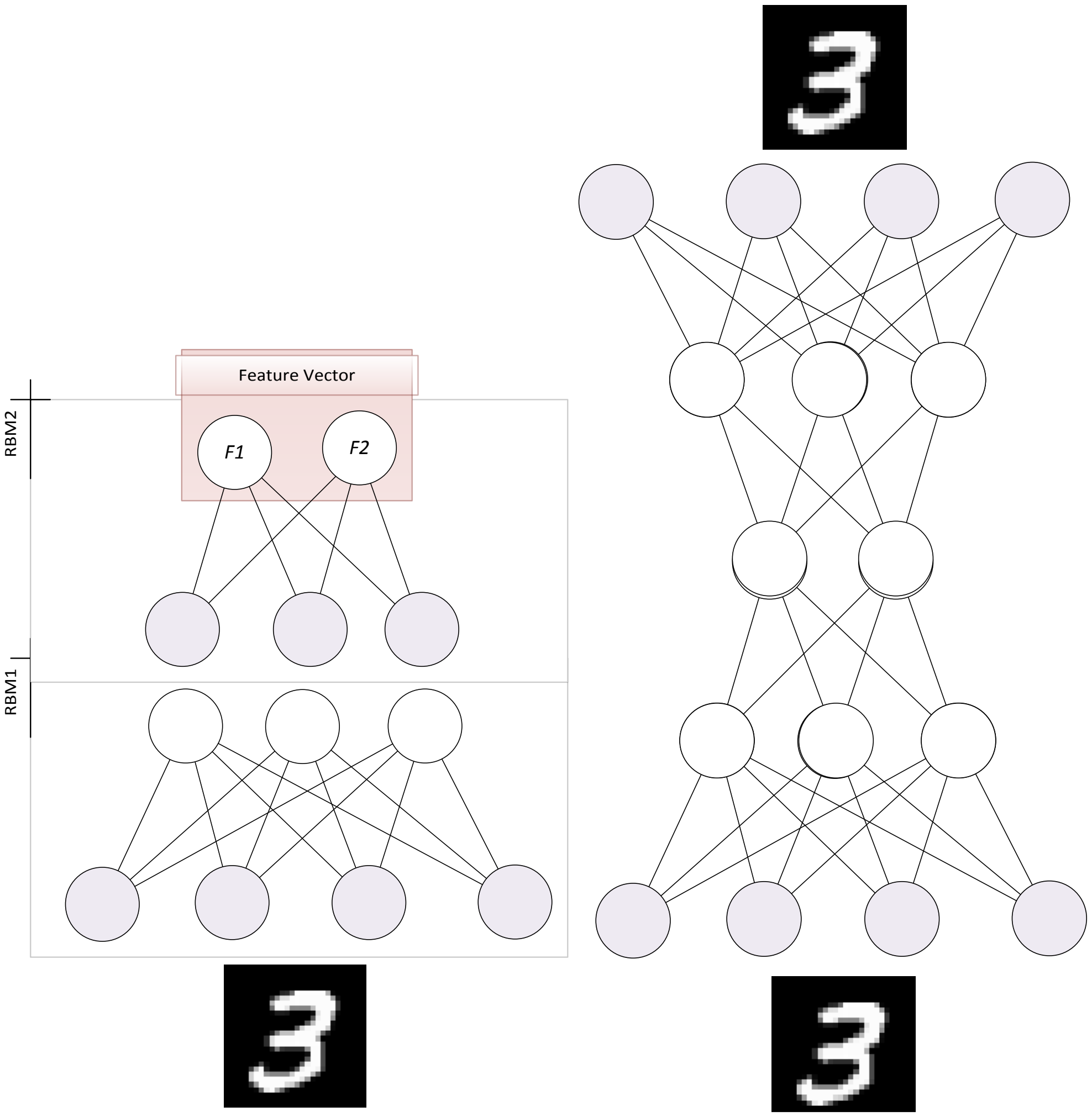

Figure 17: Conversion of an *autoEncoder* DBN to a MATLAB neural network structure. Left: A DBN with generative RBMs. Right: A neural network structure with the added upside down DBN to reconstruct input data.

The other method is *getOutput* that is used to get DBN outputs. This method returns results according to type of the DBN. Therefore in an *autoEncoder* or *classifier* DBN, results are extracted features or labels respectively.

The last method is *plotBases* that can be used to plot bases function that has been learned by DBN.

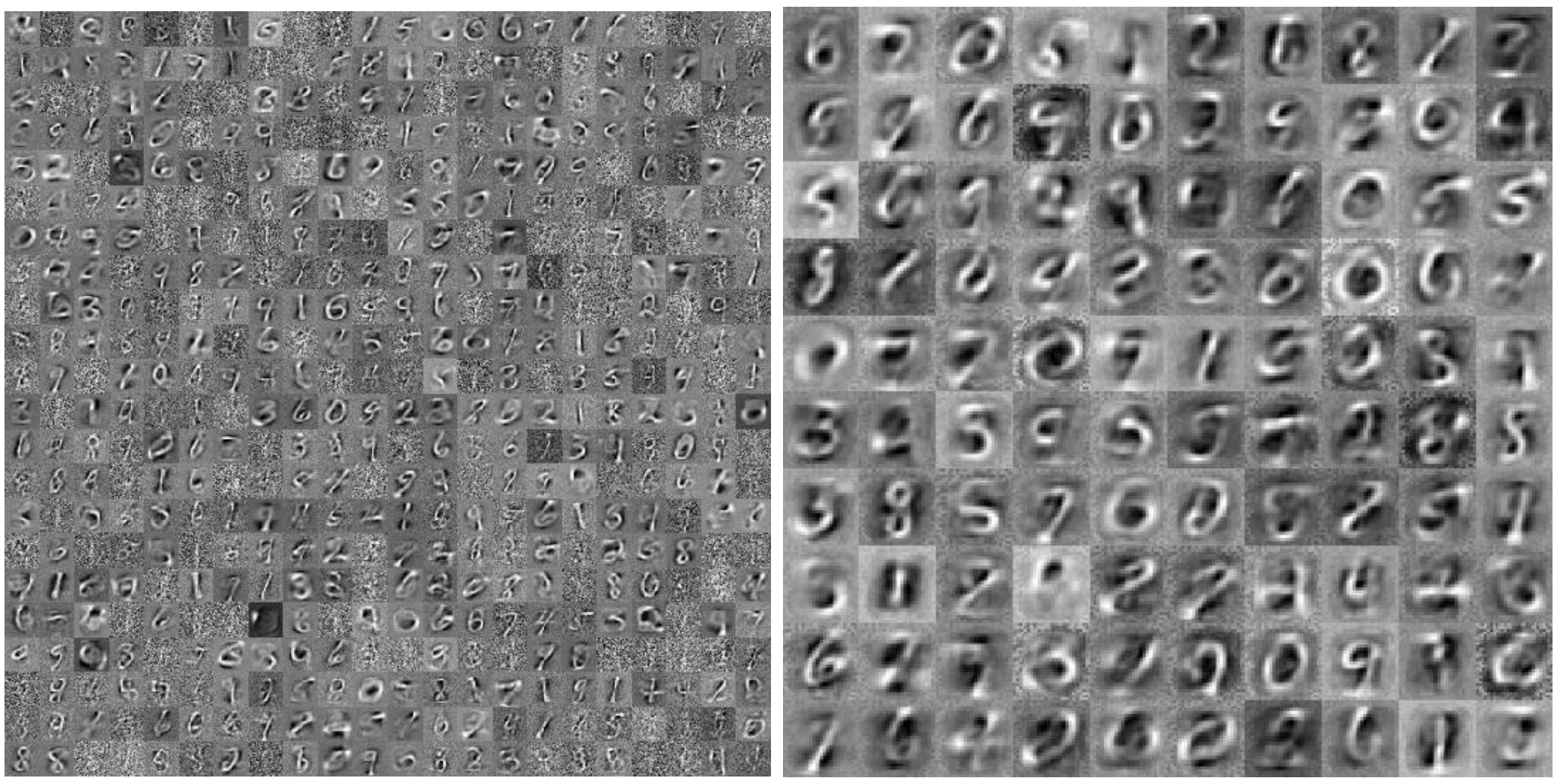

Figure 18: Bases function that *plotBases* function can plot. These bases function are from a two layer sparse normal DBN that has been learned on MNIST dataset. Left: bases function in first layer. Right: bases function in second layer.

Table 2 shows a classification experiment using this toolbox on MNIST and ISOLET dataset. This table compares different sampling method types that has been implemented in our toolbox, before and after back-propagation.

Table 2: Classification error on MNIST dataset for a DBN (784-500-500-2000) and on ISOLET dataset for a DBN (617-1000-1000-2000) and on 20 Newsgroups dataset for a DBN (5000-500-500-2000) using different sampling methods. After training each RBM, the DBN was fine-tuned in 200 epochs using back-propagation method.

| **Method** | **MNIST** | | **ISOLET** | | **20 Newsgroups** | |
|---|---|---|---|---|---|---|
| | Before BP | After BP | Before BP | After BP | Before BP | After BP |
| **CD** | 0.0636 | 0.0124 | 0.0552 | 0.0372 | 0.3087 | 0.2686 |
| **PCD** | 0.0307 | 0.0122 | 0.0500 | 0.0385 | 0.3183 | 0.2642 |
| **FEPCD** | 0.0248 | 0.0111 | 0.0449 | 0.0353 | 0.3161 | 0.2678 |

# 4. Conclusion

The paper provides a survey on the relevant literatures on DBNs and introduces a new object oriented MATLAB toolbox with most of tools necessary for conducting research and providing implementations using DBNs. In this paper some types of RBMs (such as generative or discriminative), sampling methods (such as CD, PCD and FEPCD) and DBNs (like classifier or auto encoder) have been reviewed and their implementations and a brief description of classes and methods defined in this toolkit are introduced. In addition the results of some conducted experiments using this toolkit are also presented. According to the results on MNIST (image dataset) and ISOLET (speech dataset), this toolbox can extract useful features with acceptable discrimination between them without using label information. Also on both datasets, the obtained classification performances are comparable to those reported in the state of the art literature on DBNs. In addition the toolbox can be used in other applications like generating data from trained model, reconstructing data and reducing noise.

For future work, we would like to investigate other types of RBMs and DBNs (such as convolutional DBN) and to develop our toolbox with these new types of RBMs and DBNs. Also we want to examine the toolbox using other datasets and applications and in addition we want to improve the performance.